\documentclass[letterpaper, 10 pt, conference]{ieeeconf}

\IEEEoverridecommandlockouts                              

\usepackage{cite}
\usepackage{amsmath,amssymb,amsfonts}
\usepackage{algorithmic}
\usepackage{graphicx}
\usepackage{url}
\usepackage{booktabs}
\usepackage{textcomp}
\usepackage{multicol}
\usepackage{float}
\usepackage{xcolor}

\renewcommand{\baselinestretch}{0.99}

\title{\LARGE \bf
A Large-Scale Dataset for Multi-Class Financial News Classification in Brazilian Portuguese
}

\author{Leomax Bandeira$^{1}$, Gabriel Aguiar$^{1}$, Dalton C. G. Valadares$^{1,2,*}$, Luiz Pereira$^{2}$, Helder Oliveira$^{3}$, \\ Gustavo Xavier$^{1}$, Andson M. Balieiro$^{4}$
\thanks{Authors thank 
        \textit{Coordenação de Aperfeiçoamento de Pessoal de Nível Superior (Capes), Brazil, Finance Code 001,} for partial financing of this work.}
\thanks{$^{1}$Federal University of Paraíba,
        BRA}%
\thanks{$^{2}$Federal University of Campina Grande,
        BRA}%
\thanks{$^{3}$Federal Rural University of Pernambuco,
        BRA}%
\thanks{$^{4}$Federal University of Pernambuco,
        BRA}%
\thanks{$^{*}$daltoncezane@gmail.com}%
}

\begin{document}

\maketitle
\thispagestyle{empty}
\pagestyle{empty}

\begin{abstract}
The development of Financial Natural Language Processing (NLP) systems for Brazilian Portuguese is hindered by the scarcity of domain-specific datasets. To address this, we present a large-scale corpus of approximately 160,000 Brazilian Portuguese financial news articles collected from web sources. To establish initial benchmarks for a weakly supervised multi-class classification approach based on semantic similarity, we annotated subsets of this corpus both manually and using a Large Language Model (LLM). We evaluated three similarity-based methods -- cosine similarity, K-Nearest Neighbors (KNN), and FAISS-based nearest-neighbor search -- using multilingual semantic embeddings. Our experiments reveal that the task is highly sensitive to label quality and class distribution. On the manually labeled subset, KNN achieved the best performance (47.3\% accuracy, 31.9\% F1-score), closely followed by FAISS. Conversely, on the LLM-annotated subset, while accuracy reached approximately 50\%, F1-scores dropped to near 20\%, highlighting the negative impact of label imbalance and noisy weak annotations. These findings establish foundational baselines for a challenging task, underscoring the critical need for high-quality manual annotations. Ultimately, this dataset and our preliminary results provide a reference point for future research in Brazilian Portuguese Financial NLP.
\end{abstract}

\vspace{-0.5em}
\section{INTRODUCTION}
\label{sec:Introduction}

The ability to automatically process and classify large volumes of textual data has become a central element of modern financial analysis. News articles, in particular, constitute the primary medium through which economic events, corporate developments, and information with potential market impact are disseminated to the public. Accurate and timely categorization of such content can offer significant advantages in financial decision-making, risk assessment, and market monitoring processes. Natural Language Processing (NLP) provides powerful tools for this task, and with the development of semantic representation models for Portuguese, such as BERTimbau \cite{Souza2020}, the possibilities for automated text understanding in Brazilian Portuguese have been substantially expanded.

Despite these advances, the development of financial NLP systems for Portuguese still faces a persistent obstacle: while resources such as Financial PhraseBank \cite{Malo2014} and SemEval-2017 Task 5 \cite{Cortis2017} have supported research on financial text classification and sentiment analysis in English, equivalent resources for Brazilian Portuguese financial news remain scarce.

Access to existing data represents an additional challenge. Even when Portuguese-language financial text resources exist, they are frequently difficult to obtain, restricted in scope, or limited to short time periods. This limits their representativeness and hinders their application in financial contexts, where source diversity is fundamental to model robustness. In practice, researchers interested in training or evaluating classification models for Brazilian financial news still lack an open, reproducible, and domain-appropriate reference benchmark.

The present work addresses these gaps by introducing a new large-scale dataset of Brazilian Portuguese news articles collected from multiple sources, with a focus on financial market content. The dataset comprises approximately 160,000 news articles, including 2,300 manually labeled instances organized into six domain categories: Steel Industry, Oil, Banks, Fintechs, General Market Impact, and Unrelated. From the manually labeled subset, 1,800 instances were used as a labeled reference set for training or label propagation, while 500 were reserved as a held-out test set. An additional
subset of articles was labeled by a Large Language Model
(LLM) through zero-shot prompting and used as weakly
labeled reference data. In the similarity-based classification
experiments, the held-out manually labeled test set was kept
separate and used as the ground truth for evaluation. To
avoid data leakage, test-set URLs were excluded from both
the manual and LLM-labeled reference sets before running
the classifiers. To demonstrate the utility of the dataset, we evaluated three semantic similarity-based classification approaches: Cosine Similarity, K-Nearest Neighbors (KNN), and FAISS-based nearest-neighbor search, using multilingual sentence embeddings. The results establish baseline references for future work in Financial NLP for Brazilian Portuguese.

The remainder of this paper is organized as follows. Section II reviews related work. Section III presents the dataset, its structure, and the adopted labeling scheme. Section IV describes the baseline experiments and methodology. Section V discusses the experimental results, and Section VI concludes the paper with directions for future research.

\vspace{-0.5em}
\section{RELATED WORKS}
\label{sec:Related_Works}

One important challenge in large-scale text classification is the presence of noise introduced during data collection and text extraction. Vinciarelli \cite{Vinciarelli2004} shows that noisy textual features can reduce classification robustness, especially in traditional classifiers such as Support Vector Machines. This issue is particularly relevant in large-scale web collection scenarios, where automatically retrieved documents may contain irrelevant or weakly related content.

Several benchmark datasets have supported the development of Financial NLP, especially in English. Financial PhraseBank provides human-annotated financial sentences for sentiment analysis and has become a standard resource for evaluating financial text classification methods \cite{Malo2014}. SemEval-2017 Task 5 introduced a benchmark for fine-grained sentiment analysis in financial microblogs and news headlines, with tracks covering both microblog messages and news statements/headlines \cite{Cortis2017}. These resources demonstrate the importance of domain-specific evaluation data, but they focus mainly on English and sentiment-oriented tasks rather than sector-level classification of Brazilian financial news.

In addition, recent studies have explored the incorporation of external knowledge and contextual information into financial prediction models. Ahn et al. \cite{Ahn2025} introduce a Bayesian ensemble framework that performs dynamic variable selection over macroeconomic indicators to detect financial events such as market stress and inflation. Such approaches highlight the importance of modeling external financial events, yet they typically operate under the assumption that input data are already curated and relevant to the target domain.

Semantic embeddings have become a common representation for text similarity and retrieval tasks. Sentence-BERT introduced sentence-level embeddings that can be compared using cosine similarity, reducing the computational cost of semantic similarity search compared to standard BERT pairwise inference \cite{Reimers2019}. In weakly supervised settings, such representations allow a small labeled reference set to be used for label propagation through nearest-neighbor search in the embedding space. This is particularly valuable when manual annotation is expensive or when working with languages that lack large annotated corpora.

Similarity-based classification can be implemented through exact or approximate nearest-neighbor methods. K-Nearest Neighbors is a classical non-parametric classifier that assigns labels based on the closest labeled examples in the feature space \cite{Cover1967}. In large embedding collections, exact nearest-neighbor search becomes computationally expensive. FAISS addresses this limitation by providing efficient indexing structures for large-scale similarity search \cite{Johnson2019}. Hashing-based methods such as SimHash offer another scalable alternative by representing documents as compact binary signatures and comparing them through Hamming distance \cite{Charikar2002}. These approximate methods are essential for deploying similarity-based classification in production environments with large document collections.

Finally, corpus-based approaches remain relevant for understanding semantic patterns in large-scale textual data. Mason \cite{Mason2004} introduces a system for extracting metaphorical mappings from large corpora, demonstrating that statistical analysis of linguistic patterns can uncover semantic relationships between domains. Although not specific to finance, this work highlights the importance of large-scale corpus analysis for extracting structured knowledge from text, which is relevant for financial text mining tasks.

Overall, the literature demonstrates a clear trend toward combining textual data, structured financial data, and advanced machine learning techniques to better capture the complexity of financial markets. However, many studies focus primarily on prediction tasks and assume the availability of clean and structured textual datasets. In practice, large-scale text collection from web sources introduces significant noise, requiring automatic filtering and dataset construction methods before the data can be used in downstream financial NLP tasks. Methods for large-scale financial text filtering, weak supervision, and semantic similarity-based classification thus become essential steps in the construction of financial text datasets, particularly for languages with limited labeled resources such as Brazilian Portuguese. Despite these advances, most existing work assumes the availability of clean and well-structured datasets, which is unrealistic in large-scale web data collection scenarios. Furthermore, few studies focus on dataset construction and weak supervision strategies for languages other than English, leaving a gap that this work aims to address by proposing a practical framework for filtering and labeling Brazilian Portuguese financial news at scale.

\vspace{-1em}
\section{Dataset Description}
\label{sec:Dataset_Description}

This section describes the dataset used in this work, covering the news collection process, the dataset structure, and the adopted labeling scheme.

\subsection{Data Scraping}
\label{subsec:Data_Scraping}

The dataset was built from the automated collection of Brazilian Portuguese news articles using search terms related to financial market segments of interest: \textit{Steel Industry}, \textit{Oil}, \textit{Bank}, \textit{Fintech}, and \textit{General Market Impact}. For each retrieved article, the original URL and publication date were stored. The collection covers the period from January 1, 2020 to December 30, 2025, comprising approximately 160,000 news articles.

The collection strategy was intentionally broad, since search-based retrieval often returns documents that are only partially related, weakly related, or unrelated to the target financial domain. This characteristic reflects a realistic large-scale web collection scenario and motivates the inclusion of an \textit{Unrelated} category (i.e., out-of-domain) in the labeling scheme. Furthermore, it allows the dataset to support both filtering and classification tasks, making it suitable for downstream financial NLP applications that require prior data curation.

\begin{table*}[ht!]
\centering
    \caption{Description of dataset columns for financial news classification}
    \vspace{-1em}
    \label{tab:dataset-columns}
\begin{tabular}{@{}ccc@{}}
\toprule

        \multicolumn{1}{c}{\textbf{Column}} &
        \multicolumn{1}{c}{\textbf{Type}} &
        \multicolumn{1}{c}{\textbf{Description}} \\\hline\hline

        \texttt{url} & string & Original URL of the news article collected from the web \\\hline
        \texttt{date} & date & Publication date of the news article \\\hline
        \texttt{text} & string & Textual content used for embedding generation and classification \\\hline
        \texttt{label\_manual} & string & Category manually assigned by a human annotator \\\hline
        \texttt{label\_ai} & string & Category assigned by a large language model (LLM) via \textit{prompt} \\\hline
        \texttt{embedding} & vector &  Multilingual semantic embedding used by the similarity-based methods. \\\hline

    \end{tabular}%

\vspace{-2em}
\end{table*}

The search terms were defined in Portuguese and derived from relevant segments of the Brazilian financial market. They include company names, tickers traded on B3, and expressions frequently associated with the selected sectors, ensuring coverage at both the corporate and macroeconomic levels. The collection process was implemented using \textit{Selenium} and \textit{BeautifulSoup} \cite{Mitchell2018}. \textit{Selenium} was used to automate navigation on pages with dynamic content, while \textit{BeautifulSoup} was used to parse the returned HTML and extract the fields of interest.

\subsection{Dataset Structure}
\label{subsec:Dataset_Structure}

The dataset was organized in Apache Parquet\footnote{https://parquet.apache.org/} format, as described in Table~\ref{tab:dataset-columns}. Each instance corresponds to a news article. The \texttt{url} field serves as the unique identifier and enables traceability to the original source. The main labeling columns are \texttt{label\_manual}, assigned by human annotators, and \texttt{label\_ai}, assigned by an LLM for the held-out test subset. For public release, the dataset is intended to prioritize reproducibility through text, data, URLs and embeddings, while respecting copyright restrictions associated with the redistribution of full news text. For this reason, the full text of the news articles was not included; only the first 100 characters of each article are provided. However, the complete text can be obtained from the corresponding links, which does not compromise the reproducibility of the experiments and provides a means for using the dataset in its entirety.

The manually labeled subset contains 2,300 instances. From this subset, 1,800 instances were used as the labeled reference set for training or label propagation, while 500 instances were reserved as a held-out test set. 
The predictions produced by cosine similarity, KNN, and FAISS are treated as experimental outputs and are discussed in Section~\ref{sec:Baseline_Experiments}, rather than as primary dataset columns.

\subsection{Label Schema}
\label{subsec:Label_Schema}

News articles are classified into six mutually exclusive categories: \textit{Steel Industry}, \textit{Oil}, \textit{Bank}, \textit{Fintech}, \textit{General Market Impact}, and \textit{Unrelated}. The first four categories correspond to specific financial market sectors. The \textit{General Market Impact} category captures news with potential cross-sector effects, such as macroeconomic announcements, regulatory changes, or major political and geopolitical events. The \textit{Unrelated} category includes articles retrieved during the collection process but not directly related to the investigated financial domain.

Manual labels were assigned by human annotators and are used as the ground truth for evaluation. However, part of the manual annotation process was supported by automated keyword searches. This strategy was adopted because the original corpus presented a strong overrepresentation of general-market and out-of-domain articles. Therefore, automated searches were used to retrieve candidate articles associated with underrepresented sectors and to build a more diverse labeled reference set. These automated procedures were used only to support candidate selection; the final category assignment remained manual.

LLM-based labels were generated through zero-shot prompting with Llama 3.2\footnote{https://ollama.com/library/llama3.2} for an additional subset of articles and used as weak annotations in an alternative reference-labeling scenario. Unlike the manually labeled subset, the articles submitted to LLM labeling were selected arbitrarily from the corpus. As a result, differences in the selection strategy may have contributed to the discrepancy in class proportions observed between the manual and LLM-based labeling distributions. In particular, the manual subset reflects a curated annotation process designed to improve class diversity, while the LLM-labeled subset reflects a less controlled selection of instances.

This design allows the evaluation to measure both the performance of similarity-based classifiers and the effect of using LLM-generated labels instead of human annotations, providing insights into the trade-offs between manual and automated labeling strategies.

\section{BASELINE EXPERIMENTS}
\label{sec:Baseline_Experiments}

The news collection process resulted in a large corpus with approximately 160,000 articles obtained through automated web scraping using search queries related to the financial sector. However, search-based collection strategies often retrieve heterogeneous documents, including articles that may be only partially related or unrelated to the original search term. This problem is common in large-scale news mining pipelines and introduces noise into automatically collected datasets.

To evaluate the usefulness of the proposed dataset, we implemented a multi-class classification pipeline based on semantic similarity. The pipeline consists of four main stages: (1) definition of labeled reference and test subsets, (2) embedding generation, (3) similarity-based classification, and (4) evaluation against human annotations. The objective is to assess how well similarity-based methods can classify Brazilian Portuguese financial news into the predefined sector-level categories.

\subsection{Experimental Setup}

A total of 2,300 articles were manually annotated according to the label schema described in Section~\ref{sec:Dataset_Description}. The manual annotation process combined random sampling with term-assisted candidate selection. Since the initial corpus was highly imbalanced, with a strong overrepresentation of general-market and out-of-domain articles, automated keyword searches were used to retrieve candidate articles associated with underrepresented sectors. These automated searches were used only to support sampling and improve the diversity of the labeled reference set; the final category assignment remained manual. This strategy was adopted to obtain a more informative training set for similarity-based classification while preserving human annotation as the ground truth.

From this manually labeled subset, 1,800 articles were used as the labeled reference set for training or label propagation, while 500 articles were reserved as a held-out test set. The test set was not used during label propagation and was used only for evaluation. In addition to the manually
labeled reference set, we evaluated an alternative weakly supervised scenario using LLM-generated labels as reference
data. These LLM-generated labels were produced through
zero-shot prompting and were not treated as ground truth. In
both the manual-reference and LLM-reference scenarios, the
held-out manually labeled test set was used for quantitative
evaluation. To avoid data leakage, URLs appearing in the
held-out test set were removed from both the manual and
LLM-labeled reference sets before running the similarity-
based classifiers. The evaluated methods were compared using accuracy and F1-score.



\subsection{Embedding Generation}

To generate semantic representations, we selected the SentenceTransformer model \texttt{paraphrase-multilingual-MiniLM-L12-v2}, which provides a balance between computational efficiency and semantic representation quality for multilingual texts. The model generates 384-dimensional embeddings. All embeddings were L2-normalized, allowing cosine similarity to be computed through the inner product between normalized vectors.

These embeddings were used as the common representation for all similarity-based classification methods evaluated in this work.

\subsection{Similarity-Based Classification}

We evaluated three similarity-based approaches: Cosine Similarity, K-Nearest Neighbors (KNN), and FAISS-based nearest-neighbor search.

The cosine similarity baseline assigns each test article the label of the most similar article in the labeled reference set. This corresponds to a one-nearest-neighbor strategy using cosine similarity over the embedding space.

The KNN method extends this approach by considering the $k$ nearest labeled examples in the embedding space and assigning the final category by majority voting among the retrieved neighbors. In this work, we set $k = 6$ as a fixed baseline configuration for nearest-neighbor classification \cite{Cover1967}.

FAISS was evaluated as an efficient nearest-neighbor retrieval strategy over the embedding space. Unlike exact KNN, which compares each query against all reference vectors, FAISS uses indexing structures to reduce retrieval cost in large-scale vector collections \cite{Johnson2019}. In this work, FAISS was used to retrieve nearest neighbors from the labeled reference set and assign labels according to the retrieved examples.




\subsection{Evaluation Protocol}

All methods were evaluated on the same held-out test set of 500 manually labeled articles. The manual annotations were used as the ground truth. We evaluated two reference-labeling scenarios. In the first scenario, similarity-based
classifiers used manually labeled articles as reference data.
In the second scenario, they used LLM-generated labels as
weakly labeled reference data. In both scenarios, predictions
were evaluated against the held-out manually labeled test set
using accuracy and F1-score. Before evaluation, URLs from
the held-out test set were excluded from the reference sets
to prevent data leakage. 


The output of this stage consists of baseline predictions and evaluation metrics for each method. These results provide reference baselines for future studies using the proposed dataset.

  \vspace{-1em}
\begin{figure}[!b]
  \begin{center}
    \includegraphics[width=1\columnwidth]{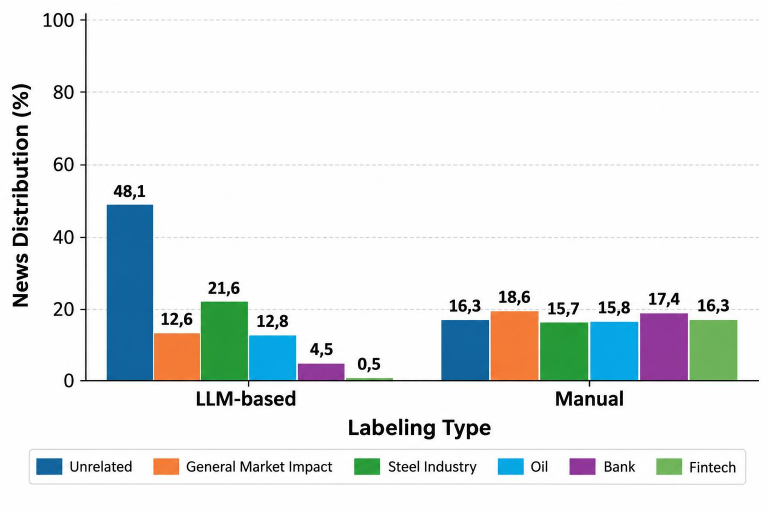}
  \end{center}
  \vspace{-1em}
  \caption{Distribution of labels under LLM-based and manual annotation.}
  
  \vspace{-1em}
  \label{fig:Distribuition_Automatic}
\end{figure}
\begin{figure*}[!th]
\begin{center}
        \includegraphics[width=1.5\columnwidth]{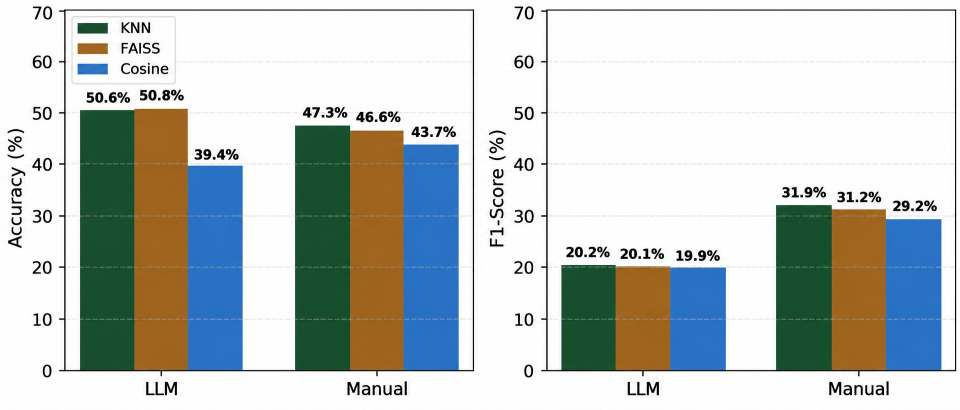}
	\end{center}
    \vspace{-1em}
	\caption{Accuracy and F1-score of similarity-based classifiers under LLM-based and manual labels.}
	\label{fig:Accuracy_Comparison}
    \vspace{-2em}
\end{figure*}

\section{RESULTS AND DISCUSSION}
\label{sec:Results_Discussions}

The construction of large-scale datasets from web sources inherently introduces noise due to the heterogeneous and unstructured nature of online content. In this study, the news collection process was based on search queries related to financial market sectors, resulting in a dataset with approximately 160,000 news articles written in Brazilian Portuguese. However, search-based collection strategies often retrieve articles that are only partially related or unrelated to the sector associated with the search query, introducing noise into the collected corpus.

To address this problem, this work evaluated a semantic similarity-based multi-class classification approach for financial news. The experiments used manually labeled articles as a reference set and a held-out manually labeled subset as the test set. In addition, an alternative scenario used LLM-generated labels as weakly labeled reference data. In both scenarios, the held-out manually labeled test set was used as ground truth, while LLM-generated labels were analyzed as weak annotations.

Figure~\ref{fig:Distribuition_Automatic} presents the distribution of news across the labeling strategies considered in the analysis (LLM-based and manual annotation settings), offering a visual comparison of how different annotation approaches distribute instances across the six categories. The LLM-generated labels are strongly concentrated in the \textit{Unrelated} category, which accounts for 48.1\% of the LLM-labeled instances. In contrast, the manually labeled subset is more evenly distributed across the six classes. This distribution reflects the annotation strategy adopted in this work: because the original corpus contained an overrepresentation of general-market and out-of-domain articles, automated keyword searches were used to retrieve candidate articles from underrepresented sectors, while the final label assignment remained manual. Therefore, the manual distribution should be interpreted as the distribution of the curated labeled subset, not as the natural distribution of the full corpus.

The news articles were represented using dense semantic embeddings, allowing semantic comparison between documents. Based on these representations, nearest-neighbor methods were applied to perform multi-class classification, assigning each test article to one of the predefined categories. The evaluated methods include cosine similarity, K-Nearest Neighbors (KNN), and FAISS-based nearest-neighbor search. Cosine similarity assigns the label of the most similar reference article, KNN assigns labels through majority voting among the nearest reference examples, and FAISS provides an efficient indexed search strategy over the embedding space. 

The results in Figure~\ref{fig:Accuracy_Comparison} show that KNN and FAISS achieved similar performance, while cosine similarity obtained lower results in both evaluation scenarios.  When using LLM-generated labels as weakly labeled reference data, FAISS achieved the highest accuracy, with 50.8\%, followed by KNN with 50.6\% and cosine similarity with 39.4\%. However, the corresponding F1-scores were much lower: 20.1\% for FAISS, 20.2\% for KNN, and 19.9\% for cosine similarity. This indicates that the apparently higher accuracy in the LLM-based setting is affected by class imbalance and does not necessarily reflect balanced classification performance across all categories.

When evaluated against manually annotated labels, the methods achieved lower accuracy but higher F1-scores. KNN obtained the best manual-label performance: 47.3\% accuracy and 31.9\% F1-score. FAISS achieved similar results (46.6\% accuracy and 31.2\% F1-score), while cosine similarity reached 43.7\% accuracy and 29.2\% F1-score. These results suggest that manual labels provide a more demanding and more balanced evaluation scenario, especially because the curated test set contains a more diverse distribution of classes.

The contrast between accuracy and F1-score is particularly important in this experiment. In the LLM-based setting, the high concentration of labels in the \textit{Unrelated} class can increase accuracy when predictions follow the dominant category, but this behavior is penalized by the F1-score. In the manual-label setting, the more balanced class distribution reduces the effect of majority-class dominance and provides a more informative evaluation of multi-class classification performance. Therefore, F1-score is especially relevant for interpreting the results of this dataset.

The low F1-score observed in the LLM-based evaluation can be explained by label noise and semantic ambiguity between financial news categories. LLMs may have difficulty distinguishing closely related categories, such as \textit{Bank}, \textit{Fintech}, and \textit{General Market Impact}, especially when articles contain broad economic terminology or indirect sectoral implications. Since similarity-based methods depend directly on the quality of the reference labels or evaluation labels, noise in LLM-generated annotations can reduce the reliability of the evaluation and affect the interpretation of model performance.

From a computational perspective, exact KNN requires comparing each query article against all articles in the reference set, which becomes costly as the number of documents increases. FAISS addresses this limitation by using indexing structures for efficient nearest-neighbor retrieval in large vector collections. Although FAISS obtained performance close to KNN in this experiment, it remains more suitable for large-scale applications because it reduces the computational cost of similarity search.

Despite the usefulness of the proposed baseline results, this work has some limitations. First, the manually annotated subset is still relatively small compared to the full corpus. Second, the manual training subset was intentionally curated to improve class diversity, which is useful for training and evaluation but may not represent the natural class distribution of the collected corpus. Third, similarity-based classification methods do not learn complex decision boundaries in the same way as supervised neural classifiers. Finally, the results depend directly on the quality of the embeddings and on the reliability of the labels used as reference.

Overall, the results demonstrate that label quality and class distribution strongly affect similarity-based financial news classification. The comparison between manual and LLM-generated labels shows that accuracy alone can be misleading under imbalanced label distributions, while F1-score provides a more informative measure of multi-class performance. The dataset produced in this work represents a new resource for Financial Natural Language Processing in Brazilian Portuguese, supporting future research on sector-level financial news classification, weak supervision, and large-scale semantic similarity methods.

Future work includes expanding the manually annotated subset, improving the validation of LLM-generated labels, investigating supervised transformer-based classifiers, and integrating financial news with structured financial data such as stock prices and economic indicators.

\section{CONCLUSIONS}
\label{sec:Conclusion}

This work investigated the use of weak supervision and semantic similarity-based methods for multi-class classification of financial news in Brazilian Portuguese using a large volume of web-collected textual data. A dataset with approximately 160,000 news articles was constructed, and a subset of 2,300 articles was manually labeled according to six financial-domain categories.  From this manually labeled subset, part of the data was used as a labeled reference set, while a held-out manually labeled test set was reserved for evaluation. An additional LLM-labeled subset was used to analyze the effect of weak annotations in similarity-based classification. Across the evaluated scenarios, manual labels in the held-out test set were used as the ground truth for quantitative evaluation. 

The experimental results showed that similarity-based methods provide useful baselines for financial news classification in Brazilian Portuguese. In the manual-label evaluation scenario, KNN achieved the best performance, with 47.3\% accuracy and 31.9\% F1-score, followed closely by FAISS, with 46.6\% accuracy and 31.2\% F1-score. Cosine similarity obtained 43.7\% accuracy and 29.2\% F1-score. These results indicate that nearest-neighbor methods based on semantic embeddings can capture relevant similarities between financial news articles, but also show that the task remains challenging due to class ambiguity and the limited size of the manually labeled subset.

The comparison with LLM-generated labels revealed an important limitation of weak annotation. Although the LLM-based scenario produced higher accuracy values for KNN and FAISS, around 50\%, the corresponding F1-scores were close to 20\%. This contrast indicates that accuracy alone can be misleading under imbalanced label distributions. The LLM-generated labels were strongly concentrated in the \textit{Unrelated} category, while the manually labeled subset was more balanced across the six classes. Therefore, the results suggest that LLM-generated labels can support annotation at scale, but they require careful validation before being used as reliable ground truth in financial NLP tasks.

FAISS achieved performance close to exact KNN while offering a more scalable nearest-neighbor retrieval strategy. This makes FAISS particularly suitable for large-scale applications involving dense vector representations and large document collections. 

The main contributions of this work include the construction of a large-scale Brazilian Portuguese financial news corpus, the creation of a manually labeled subset with a sector-level taxonomy, the comparison between manual and LLM-generated labels, and the evaluation of semantic similarity-based classification methods (cosine similarity, KNN, and FAISS). The results highlight that label quality and class distribution are central factors in weakly supervised financial text classification.

Future work includes: expanding the manually annotated dataset; improving the validation of LLM-generated labels; investigating supervised transformer-based models for direct text classification; and integrating financial news with structured financial data such as stock prices and economic indicators, enabling applications in market prediction, event detection, and financial monitoring.

\addtolength{\textheight}{-12cm}   






\bibliographystyle{ieeetr}
\bibliography{refs}

\end{document}